\documentclass{article}
\usepackage{spconf,amsmath,graphicx}
\usepackage{graphicx}
\usepackage{algorithm}  
\usepackage{algpseudocode}  
\usepackage{amsmath}  
\usepackage{cases}
\usepackage{indentfirst}
\usepackage{enumitem}
\usepackage{multirow}

\usepackage{hyperref}
\hypersetup{hypertex=true,
    colorlinks=true,
    linkcolor=green,
    anchorcolor=green,
    citecolor=green}

\title{BBA-net: A bi-branch attention network for crowd counting}
\name{Yi Hou$ ^{1}$, Chengyang Li$ ^{2}$, Fan Yang$ ^{1}$, Cong Ma$ ^{1}$, Liping Zhu$ ^{2}$, Yuan Li$ ^{1,\dagger}$, Huizhu Jia$ ^{1}$, Xiaodong Xie$ ^{1}$
    \thanks{This work is partially supported by the National Key Research and Development Program of China under contract No. 2016YFB0401904 and the National Science Foundation of China under contract No. 61971047. $ ^{\dagger}  $This author is the corresponding author. }
}
\address{$ ^{1} $National Engineering Laboratory for Video Technology, Peking University\\
    $ ^{2} $Key Lab of Petroleum Data Mining, China University of Petroleum (Beijing)\\
    \{yihou, fyang.eecs, Cong-Reeshard.Ma, yuanli, hzjia, donxie\}@pku.edu.cn\\
    lcylmhlcy@gmail.com, zhuliping@cup.edu.cn}

\begin{document}
    %
    \maketitle
    \begin{abstract}
        In the field of crowd counting, the current mainstream CNN-based regression methods simply extract the density information of pedestrians without finding the position of each person. This makes the output of the network often found to contain incorrect responses, which may erroneously estimate the total number and not conducive to the interpretation of the algorithm. To this end, we propose a Bi-Branch Attention Network (BBA-NET) for crowd counting, which has three innovation points. i) A two-branch architecture is used to estimate the density information and location information separately. ii) Attention mechanism is used to facilitate feature extraction, which can reduce false responses.    iii) A new density map generation method combining geometric adaptation and Voronoi split is introduced. Our method can integrate the pedestrian's head and body information to enhance the feature expression ability of the density map. Extensive experiments performed on two public datasets show that our method achieves a lower crowd counting error compared to other state-of-the-art methods.
    \end{abstract}


    \section{Introduction}
    Crowd counting plays an important role in public safety, especially in crowd scenes such as concerts, sporting events, and celebrations. Without proper management, stomping events can occur and the basic premise of management is to count people. Crowd counting is a challenging task due to the occlusions, perspective distortions, and person distributions. Many researches have been done to solve these problems. Among them, the CNN-based regression methods take a crowd image as input and produce a density map which is further accumulated to get the number of people. The head sizes in the same image may greatly vary, which poses a problem for CNN to extract scale-invariant features. Many methods focus on dealing with this problem, including multi-column networks, scale aggregation modules, and scale invariance architectures. The typical multi-column methods include multi-column fusion \cite{Zhang2016Single} and deep-shallow network fusion \cite{Boominathan2016CrowdNetAD}.
    The typical scale aggregation modules \cite{cao2018scale,zeng2017multi} aggregate scale-invariant features by different kernel sizes. The typical scale invariance architectures \cite{li2018csrnet,wang2018defense,huang2018stacked,Hossain2019CrowdCU,hossain2019crowd,sindagi2019ha} focus on the design of single-column architectures. Other works also explore the weak supervision \cite{Liu2018LeveragingUD,liu2019exploiting,sam2019almost} to leverage unlabeled data.
    
    \begin{figure}[]
        \centering
        \includegraphics[width=1\linewidth]{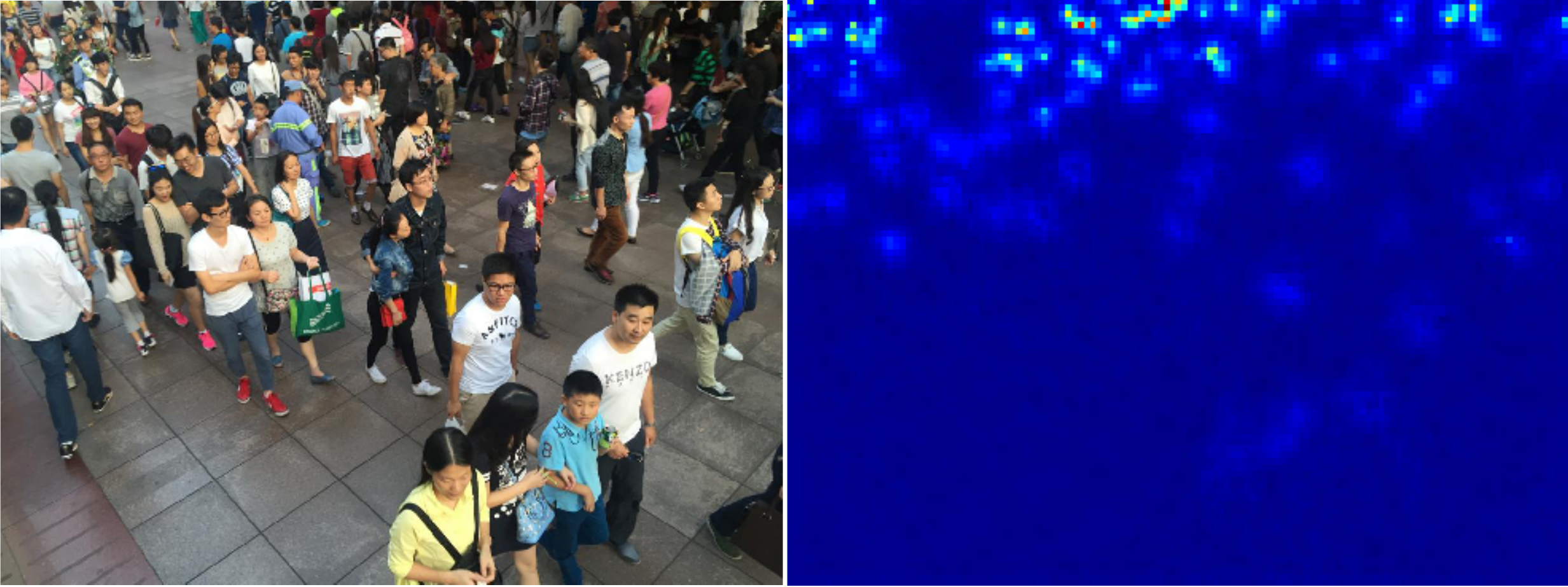} 
        \caption{A crowd image with its corresponding density map.} 
        \label{example}
    \end{figure}
    \begin{figure*}[htb]
        \centering
        \includegraphics[width=1\linewidth]{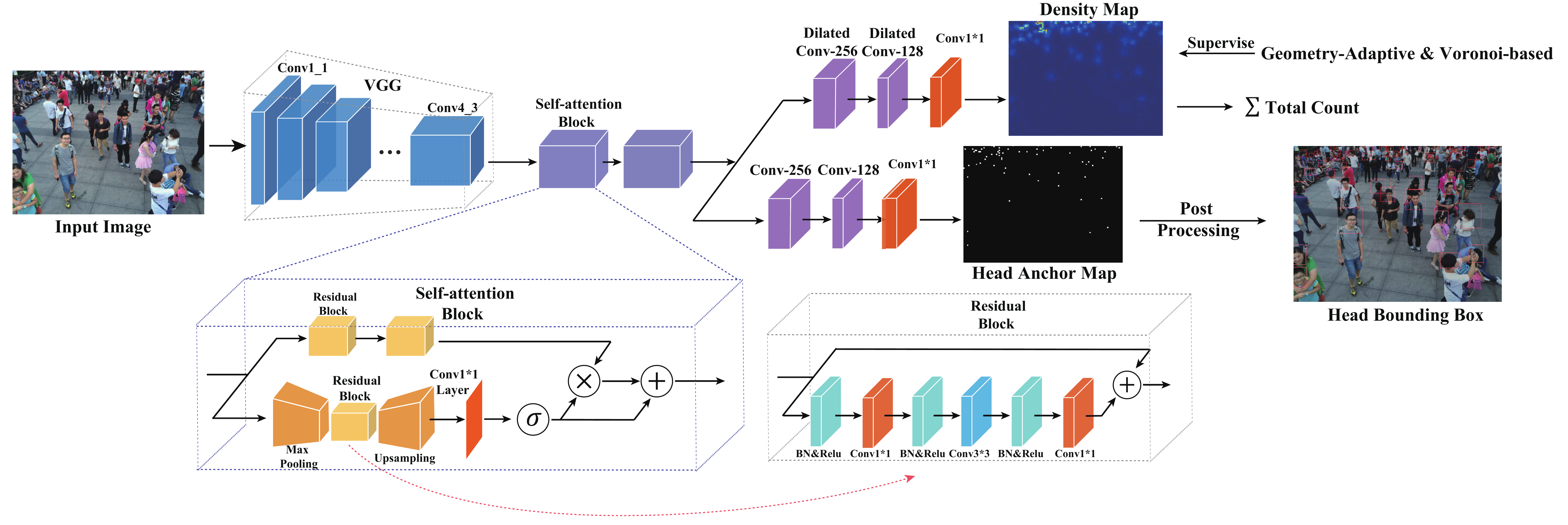} 
        \caption{The architecture of the proposed method.}
        \label{network}
    \end{figure*}

    We focus on resolving two challenges in this paper: 1) The previous crowd counting algorithms \cite{Zhang2016Single, Sam2017Switching, Boominathan2016CrowdNetAD, cao2018scale} only predict the density map, which cannot see the heads been successfully or unsuccessfully detected. Thus, the performance of the network cannot be explained. For example, which factor contributes the most to total losses?  To tackle this problem, we innovatively design a bi-branch network via adding a head location map (also head anchor map) prediction branch. The predicted head anchor map is sparse so we can get head detection boxes by post-processing. Besides, we propose to use the attention mechanism to refine pedestrian features, which can reduce the error responses shown in the network's output. 2) Most of the previous crowd counting algorithms \cite{cao2018scale, Boominathan2016CrowdNetAD, Zhang2016Single} adopt a geometry-adaptive method to produce density maps. We find that using such a method, the network can misrecognize traffic signs and billboards as pedestrian's heads.
    Because the network can only learn pedestrian's head features without context such as body features. To this end, we exploit a Voronoi-based method to produce context-awareness density maps. By combining the geometry-adaptive method and Voronoi-based method, our experiments show that this strategy can increase effectiveness and robustness.
    
    We evaluate our method on two crowd datasets ShanghaiTech and UCF\_CC\_50. The results show that our method can consistently outperform state-of-the-art methods.

    \section{Proposed Method}
    \subsection{Voronoi-Based Density map}
    Most of the previous methods adopt a geometry-adaptive algorithm \cite{Zhang2016Single} to produce the ground truth density map $ F_{geo} $,
    
    \begin{equation}\label{key}
        \setlength{\abovedisplayskip}{1pt}
        \setlength{\belowdisplayskip}{1pt}
        F_{geo}(x)=\sum_{i=1}^{N}{\delta(x-x_{i})*G_{\delta_{i}}(x)}
    \end{equation}
    where  $ \delta_{i}=\beta \overline {d}_{i} $, and $ \overline {d}_{i} = \frac{1}{m}\sum_{j=1}^{m}d_{j}^{i}$. $ \overline {d}_{i} $ represents the average distance from the annotated point to $ m $ nearest neighbors. This method ignores the pedestrian's body features, which may affect the high-quality feature extraction and density map regression. The misrecognition of traffic signs and billboards are due to it. To resolve this problem, we propose a new density map generation method based on Voronoi split. 
    
    \begin{figure}[htb]
        \centering
        \includegraphics[width=1\linewidth]{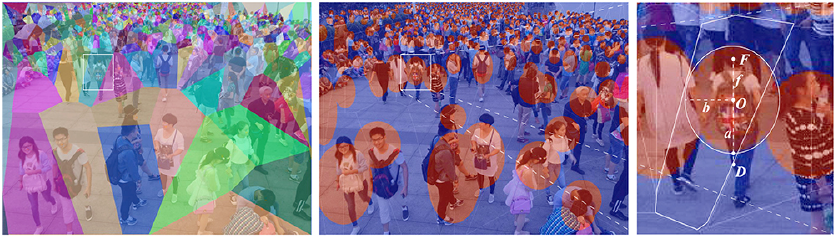}
        \caption{The left figure is a demo of Voronoi split and the middle figure shows the Gaussian kernel considering the body contexts of pedestrians.}
        \label{voronoi}
    \end{figure}
    As shown in Fig. \ref{voronoi}, Voronoi split is operated on all annotated points (head anchors) in the images. After splitting, each pixel is affiliated to the closest annotated point. Then, we draw a vertical line from the head anchor $ F $ downward and intersect the corresponding Voronoi polygon at $ D $. An ellipse is constructed inside the polygon, whose semi-axis $ a $ and $ b $ can be estimated by:
    \begin{equation}\label{key}
        \setlength{\abovedisplayskip}{1pt}
        \setlength{\belowdisplayskip}{1pt}
        f+a =\gamma d \ \ , \ b=\overline{l}
    \end{equation}
    where $ f $ indicates the focal length, $ d $ denotes the distance between $ F  $  and $ D $,  $ \overline{l}$ represents the average distance from the annotated point to $ k $ nearest Voronoi polygon edges, and  $ \gamma $ is an empirical parameter and we set $ \gamma=0.8 $. We take $ a $ and $ b $ as standard deviations of Gaussian kernel, so the proposed density map is given by

    \begin{equation}\label{key}
        \setlength{\abovedisplayskip}{1pt}
        \setlength{\belowdisplayskip}{1pt}
        F_{vor}(x)=\sum_{i=1}^{N}{\delta(x-x_{i})*G_{\delta_{i}^1,\delta_{i}^2}(x)}
    \end{equation}where  $ \delta_{i}^1=\eta a$ , $ \delta_{i}^2=\eta b$, and we set the empirical parameter $ \eta =1 $.
    
    Due to the pedestrian's body block one another in highly crowded scenes, the head may be the only source for feature extraction.  The above-mentioned methods are combined to produce the final density map.
    \begin{equation}\label{key}
        F=(1-\lambda) F_{geo}+\lambda F_{vor}
    \end{equation}where $ \lambda$ indicates the weight of two density maps, and we set $ \lambda $ is set to 0.5 in the following experiments.

    \subsection{Network Architecture}
    As shown in Fig. \ref{network}, the designed architecture consists of three stages: 1) feature extraction that aims to learn low-level features, 2) self-attention that aims to refine the learned features and 3) map regression that aims to learn anchor map and density map simultaneously.
    
    The feature extraction module is constructed by stacking the first 13 layers of VGG16 \cite{Simonyan2015VeryDC}, which contains three max-pooling layers.  The self-attention module \cite{YANG2019143} is adopted to generate attention map, targeting to attend to pedestrian relative features. In particular, features from the feature extraction module are fed into a self-attention module which consists of two paths, one for high-level feature extraction and the other for attention estimation. The feature extraction path is built by two residual blocks \cite{He2016DeepRL} that act as multiple detectors to extract semantic structures. Inspired by the attention mechanism that coherently understands the whole image and further focus on local discriminative regions, we adopt attention blocks here. The attention path is built by an encoder-decoder network that acts as a mask function to re-weight the features for automatic inference of regions of interest. It is difficult to directly learn anchor map from the extracted feature since the anchor map is more discrete than the density map. On the other hand, density map regression can provide common information for high-quality anchor map learning. Moreover, it is easy to converge, due to the smoother Gaussian filter. Inspired by this observation, we design a novel map regression strategy, which decomposes regression into a density map regression component and an anchor regression component. The decomposition facilitates the optimization significantly without bringing extra complexity in inference.
    
    We use Euclidean distance as the loss function for both two branches, which can be defined by:
    {\setlength\abovedisplayskip{1pt}
        \setlength\belowdisplayskip{0.5pt}
        \begin{equation}\label{key}
            L_{den}(\Phi)=\frac{1}{2N}\sum_{i=1}^{N}{\| F^{den}(X_{i};\Phi)-F^{den}_{i} \|_{2}^{2}}
    \end{equation}}
    {\setlength\abovedisplayskip{1pt}
        \setlength\belowdisplayskip{1pt}
        \begin{equation}\label{key}
            L_{anc}(\Phi)=\frac{1}{2N}\sum_{i=1}^{N}{\| F^{anc}(X_{i};\Phi)-F^{anc}_{i} \|_{2}^{2}}
    \end{equation}}where $ \Phi $ is a set of learnable parameters in our model, $ N $ is the number of training image, $ X_{i} $is the input image. $ F^{den}_{i} $ and  $ F^{anc}_{i} $ are the ground truth. $ F^{den}(X_{i};\Phi) $ stands for the estimated density map, and $ F^{anc}(X_{i};\Phi) $ stands for the estimated anchor map.
    The final combinatorial loss $ L $ is formulated as a weighted sum of $ L_{den} $ and $ L_{anc} $ as:

    \begin{equation}\label{key}
        \setlength\belowdisplayskip{1pt}
        \setlength\abovedisplayskip{1pt}
        L=\omega L_{den}+(1-\omega) L_{anc}
    \end{equation}
    where $ \omega $ is a factor to balance the contributions of $ L_{den} $ and $ L_{anc} $. We empirically set $ \omega =0.5 $ in our experiments.

    \subsection{Training Details and Post Processing}
    
    We train our model in an end-to-end manner from the VGG16 pre-trained model. Moreover, we optimize the network parameters based on the stochastic gradient descent (SGD) optimizer. The batch size is set to 50, and the initial learning rate is set to $ 10^{-6} $. The learning rate is decreased by a factor of 0.1 every 5 iterations.
    
    The produced head anchor map is more sparse than the density map as depicted in Fig. \ref{psnr_ssim}. Thus, the post-processing method can be used to roughly calculate the head detection box. The post-processing algorithm first performs a threshold to erase noise, then uses non-maximum suppression to extract head anchor. Finally, the bounding box is estimated by using the above mentioned geometry-adaptive method.

    \section{Experiments}
    
    \subsection{Evaluation Metrics}
    We use mean absolute error (MAE) and mean squared error (MSE) to evaluate our model's performance. The MAE and MSE are defined by:
    {\setlength\abovedisplayskip{1pt}\setlength\belowdisplayskip{1pt}\begin{small}
            \begin{equation}\label{key}
                MAE=\frac{1}{N}\sum_{i=1}^{N}{\vert C_{i}-C_{i}^{GT} \vert} \ \ \ \ 
                MSE=\sqrt{\frac{1}{N}\sum_{i=1}^{N}{\vert C_{i}-C_{i}^{GT} \vert}^{2}}
            \end{equation}
    \end{small}} $ C_{i} $ is the estimated number of person and $ C_{i}^{GT} $ means the ground truth number of person. In general, the MAE and MSE can show the accuracy and robustness, respectively.
    \subsection{Evaluation Results}
    \begin{figure}[tb]
        \centering
        \includegraphics[width=1\linewidth]{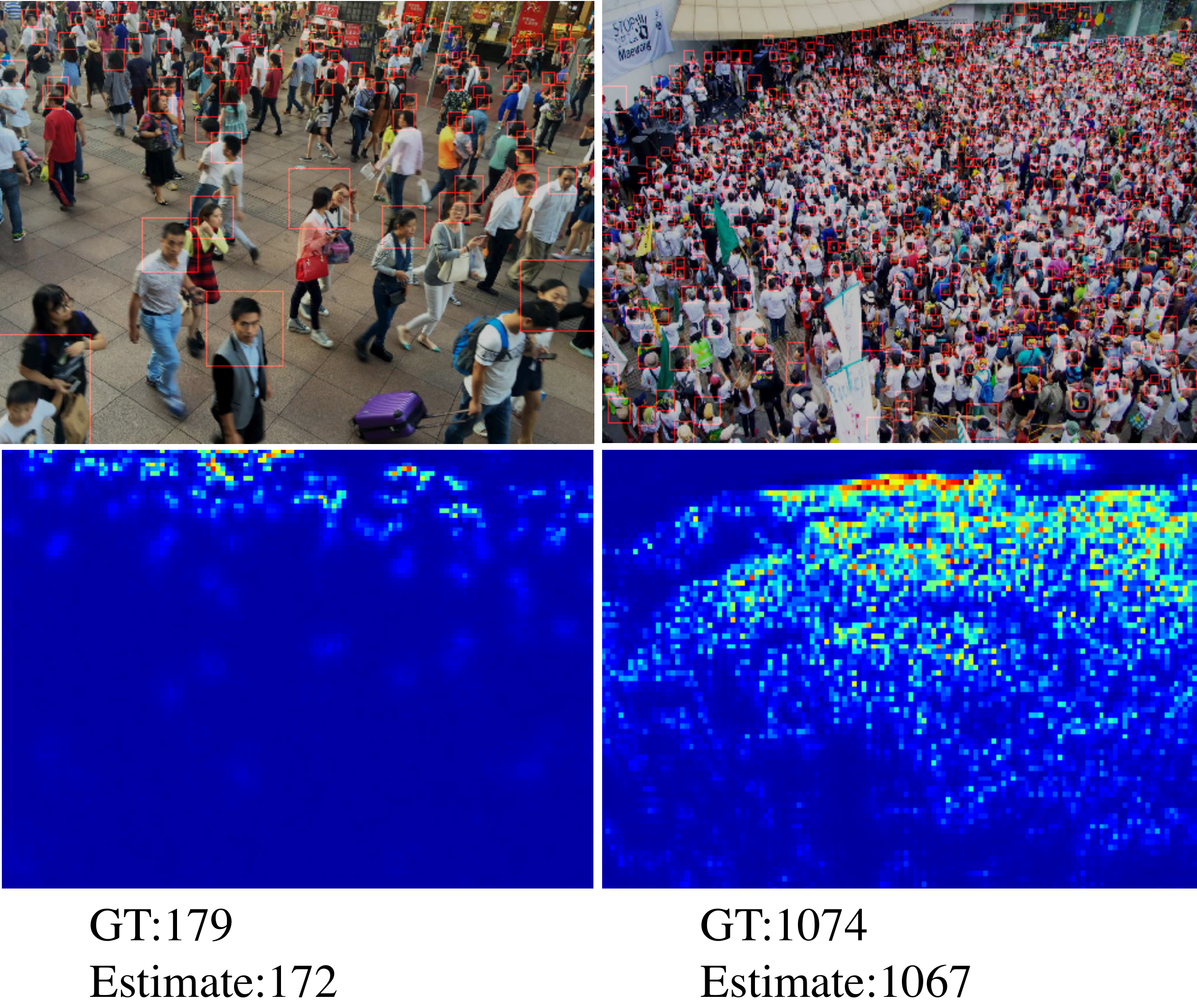} 
        \caption{Examples of the predicted head detection box and density map on the ShanghaiTech dataset. The ground truth numbers and the estimated results are also shown below.} 
        \label{figureresult}
    \end{figure}
    
    \noindent
    \textbf{ShanghaiTech.} This dataset consists of Part\_A and Part\_B which contains 1198 images, with a total number of 330,165 labeled heads. There are 482 congested images in part\_A which are downloaded from the Internet, and 716 images in Part\_B which are taken from busy streets in Shanghai. We slice the training images to patches and double them by PCA whitening. The comparison results are shown in Table \ref{result_table}, we find that our model achieves 0.4 lower MAE than the previous state-of-the-art methods in the ShanghaiTech B. Besides, our method also increases prediction robustness.

    \begin{table}[tb]
        \centering
        \caption{Comparing the performances of different methods on the ShanghaiTech dataset and UCF\_CC\_50 dataset.}
        
        \scalebox{0.95}{
            \begin{tabular}{lp{5mm}p{6mm}p{5mm}p{6mm}p{5mm}p{6mm}}
                \hline
                \multirow{2}{*}{Method} & \multicolumn{2}{c}{Part\_A} & \multicolumn{2}{c}{Part\_B} & \multicolumn{2}{c}{UCF\_CC\_50} \\ \cline{2-7} 
                & MAE & MSE & MAE & MSE & MAE & MSE \\ \hline
                Zhang et al.\cite{zhang2015cross} & 181.8 & 277.7 & 32.0 & 49.8 & 467.0 & 495.8 \\
                CP-CNN  \cite{sindagi2017generating} & 73.6 & 106.4 & 20.1 & 30.1 & 295.8 & 320.9 \\
                SaCNN  \cite{zhang2018crowd} & 86.8 & 139.2 & 16.2 & 25.8 & 314.9 & 424.8 \\
                ACSCP  \cite{shen2018crowd}& 75.7 & 102.7 & 17.2 & 27.4 & 291.0 & 404.6 \\
                IG-CNN  \cite{babu2018divide} & 72.5 & 118.2 & 13.6 & 21.1 & 291.4 & 349.4 \\
                DeepNCL  \cite{shi2018crowd} & 73.5 & 112.3 & 18.7 & 26.0 & 288.4 & 404.7 \\
                CSRNet  \cite{li2018csrnet} & 68.2 & 115.0 & 10.6 & 16.0 & 266.1 & 397.5 \\
                SANet  \cite{cao2018scale} & 67.0 & 104.5 & 8.4 & 13.6 & 258.4 & 334.9 \\
                PACNN  \cite{shi2019revisiting} & 66.3 & 106.4 & 8.9 & 13.5 & 267.9 & 357.8 \\
                HACCN \cite{sindagi2019ha} & \textbf{62.9} & 94.9 & 8.1 & 13.4 & 256.2 & 348.4 \\
                TEDnet  \cite{jiang2019crowd} & 64.2 & 109.1 & 8.2 & 12.8 & 249.4 & 354.5 \\ \hline
                Ours & 63.8 & \textbf{93.8} & \textbf{7.8} & \textbf{12.0} & \textbf{230.5} & \textbf{316.9} \\ \hline
        \end{tabular}}
        \label{result_table}
    \end{table}
    
    \noindent
    \textbf{UCF\_CC\_50.} This dataset \cite{Idrees2013Multi} contains 50 congested images of varying resolutions. The number of labeled head ranges from 94 to 4543, which is very unbalanced. We used 5-fold cross-validation to validate the performance. Considering that this dataset is so small, we use ShanghaiTech A pre-trained model to finetune it. The training just needs 5 epochs to converge. The results show that our model achieves the lowest MAE and obtains 19 lower MAE than the previous state-of-the-art methods.

    \subsection{Ablation Studies and Analysis}
    
    \noindent
    \textbf{Baseline.} Our baseline model consists of a VGG16 front-end and a density map regression branch. The density map is calculated by a geometry-adaptive method. We perform ablation studies on the ShanghaiTech B dataset. The results are shown in Table \ref{analysis}.  
    
    \noindent
    \textbf{Attention.} Attention block is a crucial module for improving performance, which aims to extract useful features. Besides, the residual units inside skip connections of the attention blocks can speed up convergence during the training phase. We also assess the effectiveness of the different number of attention blocks. The results show that we achieve the most effective performance with two attention blocks. 
    
\begin{figure}[tb]
        \centering
        \includegraphics[width=1\linewidth]{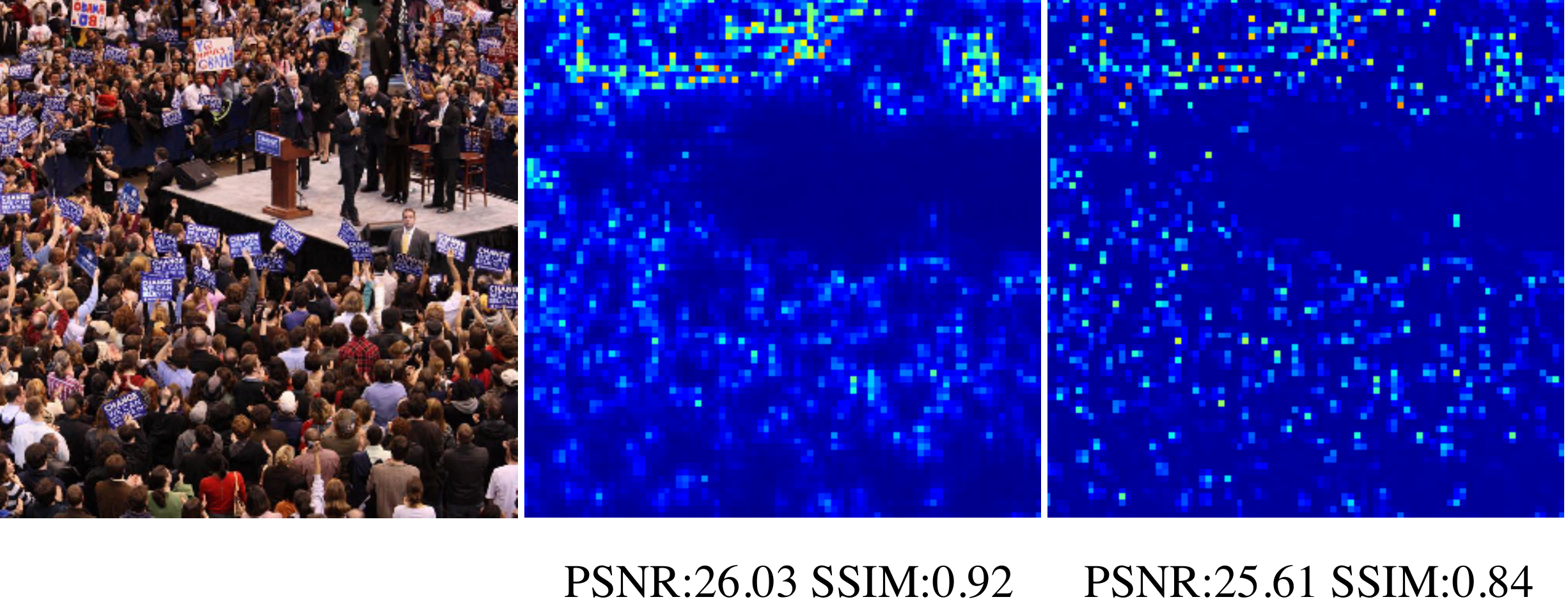} 
        \caption{The middle and right figures are the density map and anchor map produced by our network. PSNR and SSIM are also calculated below.}
        \label{psnr_ssim}
    \end{figure}

    \noindent
    \textbf{Anchor Map Branch.} We measure the gain of adding bi-branch architecture. The results show that the anchor map branch largely improves performance. We also illustrate the visualization results of the density map and anchor map using the PSNR (Peak Signal-to-Noise Ratio) and SSIM \cite{wang2004image} metrics. As shown in Fig. \ref{psnr_ssim}, the anchor map is more discrete than the density map. Due to this attribute,  the anchor map branch can add strongly prior information of head location for the density map learning, meanwhile, the density map branch promotes the anchor map branch convergence.
    
    \noindent
    \textbf{Voronoi-based Method.} We further evaluate the effect of the Voronoi-based method. As shown in Table \ref{analysis}, the performance is improved by 1.28. The results show that using the Voronoi-based method could effectively enhance performance. We also evaluate the Voronoi-based method on the UCF\_CC\_50 dataset, due to the inadequate body context information in such a crowd scene, that is the reason why it has slight improvement, so the network can only extract head features.
    
    \noindent
    \textbf{Visual Analysis.} As shown in Fig. \ref{figureresult}, we find that when people are close to another person, the network usually misrecognizes them. Furthermore, in an extremely crowded scene, the output response of the density map is locally plain. It means that strong supervision is required to extract a tiny head. It validates the utilization of our bi-branch architecture that simultaneously learning general features and head locations.
    
    \begin{table}[tb]
        \centering
        \caption{Ablation studies of self-attention, anchor map branch, Voronoi-based method on the ShanghaiTech B dataset}
        \scalebox{0.95}{
            \begin{tabular}{lp{6mm}p{6mm}}
                \hline
                Method & MAE & MSE \\ \hline
                Base & 33.8 & 50.6 \\ 
                Base+Anchor& 21.2 & 32.4 \\ 
                Base+Attention(2)& 13.1 & 18.9 \\ 
                Base+Anchor+Attention(1)& 11.0 & 17.3 \\ 
                Base+Anchor+Attention(2)& 9.2 & 13.3 \\ 
                Base+Anchor+Attention(3)& 10.6 & 16.0 \\  \hline
                Base+Anchor+Attention(2)+Voronoi&  \textbf{7.8} & \textbf{12.0} \\ \hline
        \end{tabular}}
        \label{analysis}
    \end{table}
    
    \section{Conclusion}
    In this paper, we propose a Bi-Branch Attention Network for crowd counting. The Bi-Branch architecture and attention blocks make the model extract more detailed and accurate pedestrian features. Besides, a new density map generation method based on  Voronoi is also been exploited. Extensive experiments demonstrate that our method can significantly improve the baseline model and outperforms the state-of-the-art methods.

    
    \vfill\pagebreak
    
    \small
    \bibliographystyle{IEEEbib}
    \bibliography{strings,refs}

\begin{thebibliography}{10}

\bibitem{Zhang2016Single}
Yingying Zhang, Desen Zhou, Siqin Chen, Shenghua Gao, and Ma~Yi,
\newblock ``Single-image crowd counting via multi-column convolutional neural
  network,''
\newblock in {\em Computer Vision \& Pattern Recognition}, 2016.

\bibitem{Boominathan2016CrowdNetAD}
Lokesh Boominathan, Srinivas S.~S. Kruthiventi, and R.~Venkatesh Babu,
\newblock ``Crowdnet: A deep convolutional network for dense crowd counting,''
\newblock in {\em ACM Multimedia}, 2016.

\bibitem{cao2018scale}
Xinkun Cao, Zhipeng Wang, Yanyun Zhao, and Fei Su,
\newblock ``Scale aggregation network for accurate and efficient crowd
  counting,''
\newblock in {\em Proceedings of the European Conference on Computer Vision
  (ECCV)}, 2018, pp. 734--750.

\bibitem{zeng2017multi}
Lingke Zeng, Xiangmin Xu, Bolun Cai, Suo Qiu, and Tong Zhang,
\newblock ``Multi-scale convolutional neural networks for crowd counting,''
\newblock in {\em 2017 IEEE International Conference on Image Processing
  (ICIP)}. IEEE, 2017, pp. 465--469.

\bibitem{li2018csrnet}
Yuhong Li, Xiaofan Zhang, and Deming Chen,
\newblock ``Csrnet: Dilated convolutional neural networks for understanding the
  highly congested scenes,''
\newblock in {\em Proceedings of the IEEE conference on computer vision and
  pattern recognition}, 2018, pp. 1091--1100.

\bibitem{wang2018defense}
Ze~Wang, Zehao Xiao, Kai Xie, Qiang Qiu, Xiantong Zhen, and Xianbin Cao,
\newblock ``In defense of single-column networks for crowd counting,''
\newblock {\em arXiv preprint arXiv:1808.06133}, 2018.

\bibitem{huang2018stacked}
Siyu Huang, Xi~Li, Zhi-Qi Cheng, Zhongfei Zhang, and Alexander Hauptmann,
\newblock ``Stacked pooling: Improving crowd counting by boosting scale
  invariance,''
\newblock {\em arXiv preprint arXiv:1808.07456}, 2018.

\bibitem{Hossain2019CrowdCU}
Mohammad~Asiful Hossain and Mehrdad Hosseinzadeh,
\newblock ``Crowd counting using scale-aware attention networks,''
\newblock {\em 2019 IEEE Winter Conference on Applications of Computer Vision
  (WACV)}, pp. 1280--1288, 2019.

\bibitem{hossain2019crowd}
Mohammad Hossain, Mehrdad Hosseinzadeh, Omit Chanda, and Yang Wang,
\newblock ``Crowd counting using scale-aware attention networks,''
\newblock in {\em 2019 IEEE Winter Conference on Applications of Computer
  Vision (WACV)}. IEEE, 2019, pp. 1280--1288.

\bibitem{sindagi2019ha}
Vishwanath~A Sindagi and Vishal~M Patel,
\newblock ``Ha-ccn: Hierarchical attention-based crowd counting network,''
\newblock {\em IEEE Transactions on Image Processing}, vol. 29, pp. 323--335,
  2019.

\bibitem{Liu2018LeveragingUD}
Xialei Liu, Joost van~de Weijer, and Andrew~D. Bagdanov,
\newblock ``Leveraging unlabeled data for crowd counting by learning to rank,''
\newblock {\em 2018 IEEE/CVF Conference on Computer Vision and Pattern
  Recognition}, pp. 7661--7669, 2018.

\bibitem{liu2019exploiting}
Xialei Liu, Joost Van De~Weijer, and Andrew~D Bagdanov,
\newblock ``Exploiting unlabeled data in cnns by self-supervised learning to
  rank,''
\newblock {\em IEEE transactions on pattern analysis and machine intelligence},
  2019.

\bibitem{sam2019almost}
Deepak~Babu Sam, Neeraj~N Sajjan, Himanshu Maurya, and R~Venkatesh Babu,
\newblock ``Almost unsupervised learning for dense crowd counting,''
\newblock in {\em Proceedings of the Thirty-Third AAAI Conference on Artificial
  Intelligence, Honolulu, HI, USA}, 2019, vol.~27.

\bibitem{Sam2017Switching}
Deepak~Babu Sam, Shiv Surya, and R~Venkatesh Babu,
\newblock ``Switching convolutional neural network for crowd counting,''
\newblock in {\em 2017 IEEE Conference on Computer Vision and Pattern
  Recognition (CVPR)}. IEEE, 2017, pp. 4031--4039.

\bibitem{Simonyan2015VeryDC}
Karen Simonyan and Andrew Zisserman,
\newblock ``Very deep convolutional networks for large-scale image
  recognition,''
\newblock {\em CoRR}, vol. abs/1409.1556, 2015.

\bibitem{YANG2019143}
Fan Yang, Ke~Yan, Shijian Lu, Huizhu Jia, Xiaodong Xie, and Wen Gao,
\newblock ``Attention driven person re-identification,''
\newblock {\em Pattern Recognition}, vol. 86, pp. 143 -- 155, 2019.

\bibitem{He2016DeepRL}
Kaiming He, Xiangyu Zhang, Shaoqing Ren, and Jian Sun,
\newblock ``Deep residual learning for image recognition,''
\newblock {\em 2016 IEEE Conference on Computer Vision and Pattern Recognition
  (CVPR)}, pp. 770--778, 2016.

\bibitem{zhang2015cross}
Cong Zhang, Hongsheng Li, Xiaogang Wang, and Xiaokang Yang,
\newblock ``Cross-scene crowd counting via deep convolutional neural
  networks,''
\newblock in {\em Proceedings of the IEEE conference on computer vision and
  pattern recognition}, 2015, pp. 833--841.

\bibitem{sindagi2017generating}
Vishwanath~A Sindagi and Vishal~M Patel,
\newblock ``Generating high-quality crowd density maps using contextual pyramid
  cnns,''
\newblock in {\em Proceedings of the IEEE International Conference on Computer
  Vision}, 2017, pp. 1861--1870.

\bibitem{zhang2018crowd}
Lu~Zhang, Miaojing Shi, and Qiaobo Chen,
\newblock ``Crowd counting via scale-adaptive convolutional neural network,''
\newblock in {\em 2018 IEEE Winter Conference on Applications of Computer
  Vision (WACV)}. IEEE, 2018, pp. 1113--1121.

\bibitem{shen2018crowd}
Zan Shen, Yi~Xu, Bingbing Ni, Minsi Wang, Jianguo Hu, and Xiaokang Yang,
\newblock ``Crowd counting via adversarial cross-scale consistency pursuit,''
\newblock in {\em Proceedings of the IEEE conference on computer vision and
  pattern recognition}, 2018, pp. 5245--5254.

\bibitem{babu2018divide}
Deepak Babu~Sam, Neeraj~N Sajjan, R~Venkatesh~Babu, and Mukundhan Srinivasan,
\newblock ``Divide and grow: Capturing huge diversity in crowd images with
  incrementally growing cnn,''
\newblock in {\em Proceedings of the IEEE Conference on Computer Vision and
  Pattern Recognition}, 2018, pp. 3618--3626.

\bibitem{shi2018crowd}
Zenglin Shi, Le~Zhang, Yun Liu, Xiaofeng Cao, Yangdong Ye, Ming-Ming Cheng, and
  Guoyan Zheng,
\newblock ``Crowd counting with deep negative correlation learning,''
\newblock in {\em Proceedings of the IEEE conference on computer vision and
  pattern recognition}, 2018, pp. 5382--5390.

\bibitem{shi2019revisiting}
Miaojing Shi, Zhaohui Yang, Chao Xu, and Qijun Chen,
\newblock ``Revisiting perspective information for efficient crowd counting,''
\newblock in {\em Proceedings of the IEEE Conference on Computer Vision and
  Pattern Recognition}, 2019, pp. 7279--7288.

\bibitem{jiang2019crowd}
Xiaolong Jiang, Zehao Xiao, Baochang Zhang, Xiantong Zhen, Xianbin Cao, David
  Doermann, and Ling Shao,
\newblock ``Crowd counting and density estimation by trellis encoder-decoder
  networks,''
\newblock in {\em Proceedings of the IEEE Conference on Computer Vision and
  Pattern Recognition}, 2019, pp. 6133--6142.

\bibitem{Idrees2013Multi}
Haroon Idrees, Imran Saleemi, Cody Seibert, and Mubarak Shah,
\newblock ``Multi-source multi-scale counting in extremely dense crowd
  images,''
\newblock in {\em Computer Vision \& Pattern Recognition}, 2013.

\bibitem{wang2004image}
Zhou Wang, Alan~C Bovik, Hamid~R Sheikh, Eero~P Simoncelli, et~al.,
\newblock ``Image quality assessment: from error visibility to structural
  similarity,''
\newblock {\em IEEE transactions on image processing}, vol. 13, no. 4, pp.
  600--612, 2004.

\end{thebibliography}
    
\end{document}